\documentclass[conference]{IEEEtran}
\pdfoutput=1
\IEEEoverridecommandlockouts
\usepackage{cite}
\usepackage{amsmath,amssymb,amsfonts}
\usepackage{algorithmic}
\usepackage{graphicx}
\usepackage{arydshln}
\usepackage{textcomp}
\usepackage{booktabs}
\usepackage[colorlinks,linkcolor=blue]{hyperref}
\usepackage{multirow}
\usepackage[caption=false]{subfig}
\usepackage{xcolor}
\def\BibTeX{{\rm B\kern-.05em{\sc i\kern-.025em b}\kern-.08em
    T\kern-.1667em\lower.7ex\hbox{E}\kern-.125emX}}
\newcommand{\CheckmarkBold}{\textbf{\checkmark}}
\begin{document}\sloppy

\def\x{{\mathbf x}}
\def\L{{\cal L}}

\title{HCF-Net: Hierarchical Context Fusion Network for Infrared Small Object Detection}
\author{
\IEEEauthorblockN{\textit{Shibiao Xu}$^{1}$,
\textit{ShuChen Zheng}$^{1}$, 
\textit{Wenhao Xu}$^{1}$,
\textit{Rongtao Xu}$^{3,4}$,
\textit{Changwei Wang}$^{2,3,5,}$\IEEEauthorrefmark{1},\\ \textit{Jiguang Zhang}$^{3,4}$, \textit{Xiaoqiang Teng}$^{1}$ \textit{Ao Li}$^{6,}$\IEEEauthorrefmark{1}, \textit{Li Guo}$^{2}$
}

\IEEEauthorblockA{$^{1}$Artificial Intelligence, Beijing University of Posts and Telecommunications \\ $^{2}$Key Laboratory of Computing Power Network and Information Security, Ministry of Education, \\Shandong Computer Science Center, Qilu University of Technology (Shandong Academy of Sciences) \\ $^{3}$State Key Laboratory of Multimodal Artificial Intelligence Systems, Institute of Automation, Chinese Academy of Sciences \\ $^{4}$School of Artificial Intelligence, University of Chinese Academy of Sciences \\$^{5}$ Shandong Provincial Key Laboratory of Computer Networks, Shandong Fundamental Research Center for Computer Science\\$^{6}$School of Computer Science and Technology, Harbin University of Science and Technology}

\thanks{This work is supported by Beijing Natural Science Foundation No. JQ23014, in part by the National Natural Science Foundation of China (Nos. 62271074, 62071157, 62302052, 62171321 and 62162044), and in part by the Open Project Program of State Key Laboratory of Virtual Reality Technology and Systems, Beihang University (No. VRLAB2023B01).}

\thanks{\IEEEauthorrefmark{1}Changwei Wang and Ao Li are the corresponding authors (Email: wangchangwei2019@ia.ac.cn; ao.li@hrbust.edu.cn).}

}

\maketitle
\begin{abstract}
Infrared small object detection is an important computer vision task involving the recognition and localization of tiny objects in infrared images, which usually contain only a few pixels. However, it encounters difficulties due to the diminutive size of the objects and the generally complex backgrounds in infrared images. In this paper, we propose a deep learning method, HCF-Net, that significantly improves infrared small object detection performance through multiple practical modules. Specifically, it includes the parallelized patch-aware attention (PPA) module, dimension-aware selective integration (DASI) module, and multi-dilated channel refiner (MDCR) module. The PPA module uses a multi-branch feature extraction strategy to capture feature information at different scales and levels. The DASI module enables adaptive channel selection and fusion. The MDCR module captures spatial features of different receptive field ranges through multiple depth-separable convolutional layers. Extensive experimental results on the SIRST infrared single-frame image dataset show that the proposed HCF-Net performs well, surpassing other traditional and deep learning models. Code is available at~\href{https://github.com/zhengshuchen/HCFNet}{https://github.com/zhengshuchen/HCFNet}.
\end{abstract}
\begin{IEEEkeywords}
Infrared small object detection, Deep learning, Multi-scale features.
\end{IEEEkeywords}
\section{Introduction}
\label{sec:intro}
Infrared small object detection is a crucial technology for identifying and detecting minute objects in infrared images. Due to the ability of infrared sensors to capture the infrared radiation emitted by objects, this technology enables precise detection and identification of small objects, even in dark or low-light environments. As a result, it holds significant application prospects and value in various fields, including military, security, maritime rescue, and fire monitoring.\par 
However, Infrared small object detection is still challenging for the following reasons. First, deep learning currently serves as the primary method for infrared small object detection. However, almost all existing networks adopt classic downsampling schemes. Infrared small objects, due to their small size, often come with weak thermal signals and unclear contours. There is a significant risk of information loss during multiple downsampling processes. Second, compared to visible light images, infrared images lack physical information and have lower contrast, making small objects easily submerged in complex backgrounds.\par
\begin{figure*}[t]
\begin{center}
\vspace{-0.5em}   
\setlength{\abovecaptionskip}{0cm} 
\setlength{\belowcaptionskip}{0cm} 
\includegraphics[width=0.9 \linewidth]{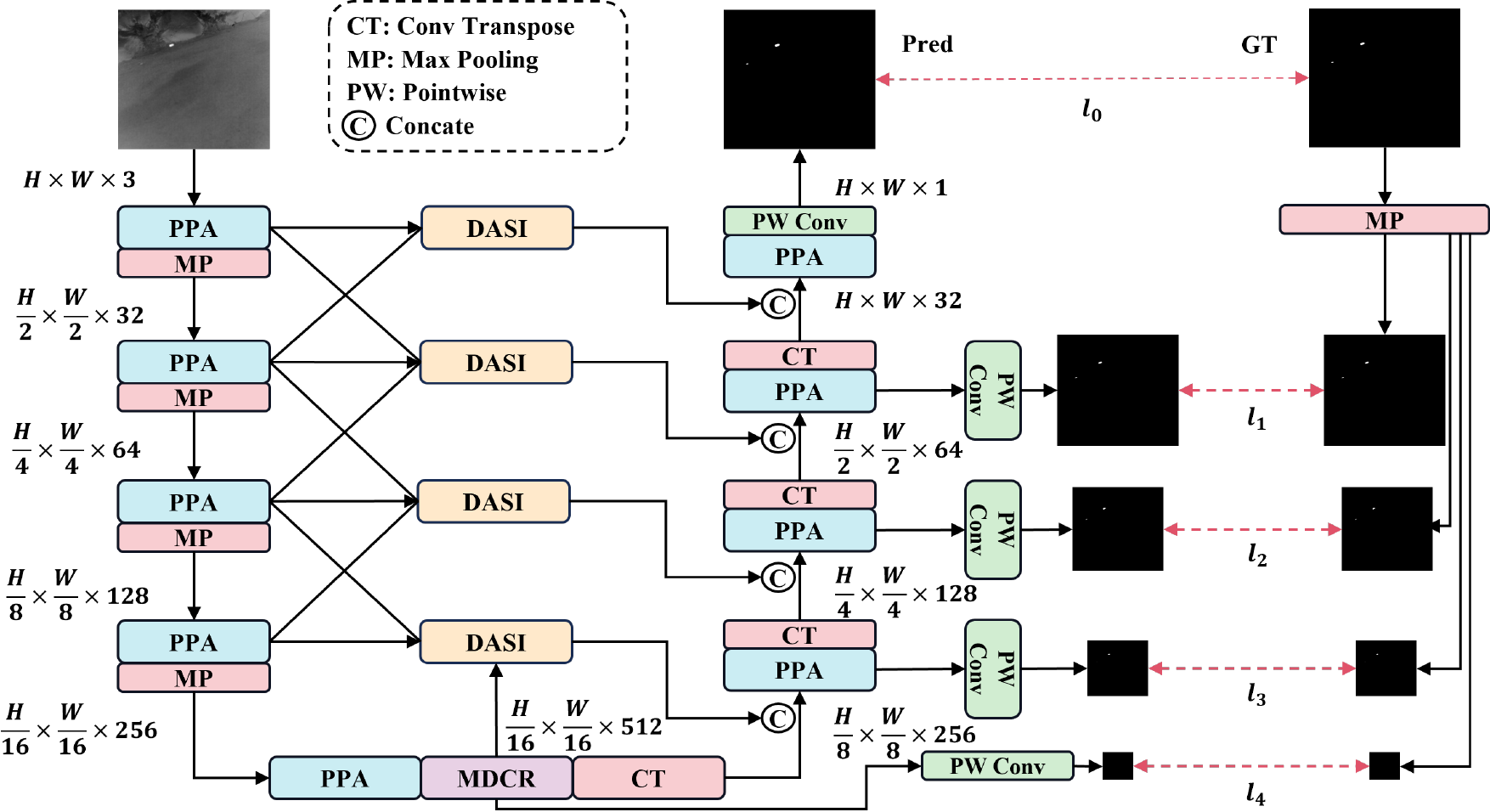}
\end{center}
   \caption{Network Architecture. The encoder primarily comprises the parallelized patch-aware attention (PPA)
   module and max-pooling layers, while the decoder mainly consists of PPA and convolutional transpose (CT) layers. We incorporate the multi-dilated channel refiner (MDCR) module as an intermediary layer to bridge the encoder and decoder. Within the skip-connection component, we introduce the dimension-aware selective integration (DASI) module to enhance the fusion and propagation of features across different network layers.
   \vspace{-0.4cm}}
\label{fig:domain_net}
\end{figure*}
To tackle these challenges, We propose an infrared small object detection model named HCF-Net. This model aims for a more precise depiction of object shape and boundaries, enhancing the accuracy of object localization and segmentation by framing infrared small object detection as a semantic segmentation problem. As illustrated in Fig.~\ref{fig:domain_net}, it incorporates three key modules: PPA, DASI, and MDCR, which address the challenges mentioned above on multiple levels.\par
Specifically, as a primary component of the encoder-decoder, PPA employs hierarchical feature fusion and attention mechanisms to maintain and enhance representations of small objects, ensuring crucial information is preserved through multiple downsampling steps. DASI enhances the skip connection in U-Net, focusing on the adaptive selection and delicate fusion of high and low-dimensional features to enhance the saliency of small objects. Positioned deep within the network, MDCR reinforces multi-scale feature extraction and channel information representation, capturing features across various receptive field ranges. It more finely models the differences between objects and backgrounds, enhancing its ability to locate small objects. The organic combination of these modules enables us to address the challenges of small object detection more effectively, improving detection performance and robustness.\par
In summary, our contributions in this paper can be summarized as follows:\par
\begin{figure*}[t]
\setlength{\abovecaptionskip}{0cm} 
\setlength{\belowcaptionskip}{0cm} 
\begin{center}
\includegraphics[width=0.9 \linewidth]{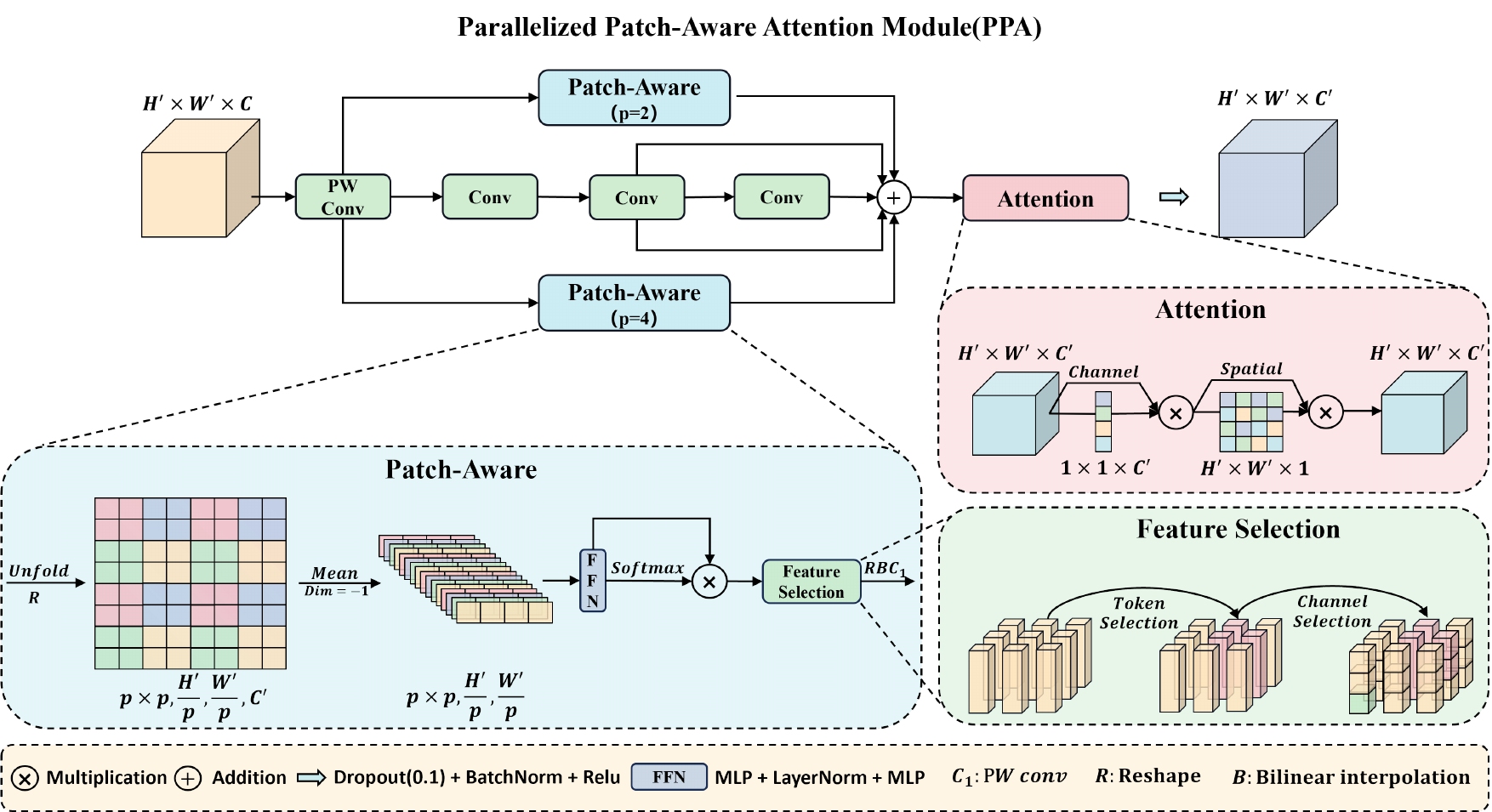}
\end{center}
   \caption{Detailed structure of the parallelized patch-aware attention module. This module primarily consists of two components: multi-branch fusion and attention mechanisms. The multi-branch fusion component includes patch-aware and concatenated convolutions. The 'p' parameter in patch-aware is set to 2 and 4, representing local and global branches, respectively.
   \vspace{-0.5cm}}
\label{fig:PPA}
\end{figure*}
\begin{itemize}
\item We model infrared small object detection as a semantic segmentation problem and propose HCF-Net, a layer-wise context fusion network that can be trained from scratch.
\item Three practical modules have been proposed: parallelized patch-aware attention (PPA) module, dimension-aware selective integration (DASI) module, and multi-dilated channel refiner (MDCR) module. These modules effectively alleviate the issues of small object loss and low background distinctiveness in infrared small object detection.
\item We evaluate the proposed HCF-Net's detection performance on the publicly available single-frame infrared image dataset SRIST and demonstrate a significant advantage over several state-of-the-art detection methods.
\end{itemize}
\section{Related Work}
\subsection{Traditional Methods}
In the early stages of infrared small object detection, the predominant approaches were model-based traditional methods, generally categorized into filter-based methods, methods based on the human visual system, and low-rank methods. Filter-based methods are typically limited to specific and uniform scenarios. For example, TopHat\cite{Zeng2006TheDO} estimates the scene background using various filters to separate the object from a complex background. Methods based on the human visual system are suitable for scenarios with large objects and strong background differentiation, such as LCM\cite{chen2013local}, which measures the contrast between the center point and its surrounding environment. Low-rank methods are suitable for fast-changing and complex backgrounds but lack real-time performance in practical applications, often requiring additional assistance such as GPU acceleration. Examples of these methods include IPI\cite{Gao2013InfraredPM}, which combines low-rank background with sparsely shaped objects using low-rank decomposition, PSTNN\cite{PSTNN} employing a non-convex method based on tensor nuclear norms, RIPT\cite{RIPT} that focuses on reweighted infrared patch tensors, and NIPPS\cite{dai2016infrared}, an advanced optimization approach that attempts to incorporate low-rank and prior constraints. While effective in specific scenarios, traditional methods are susceptible to interference from clutter and noise. In complex real-world scenarios, modeling objects is significantly affected by model hyperparameters, resulting in poor generalization performance.\par
\subsection{Deep Learning Methods}
In recent years, with the rapid development of neural networks, deep learning methods have significantly advanced the infrared small object detection task. Deep learning approaches\cite{xu2023rssformer, xu2022instance, wang2022cndesc, wang2022net, wang2022mtldesc, wang2023attention, wang2023triple, xu2023self}  exhibit higher recognition accuracy than traditional methods without relying on specific scenes or devices, demonstrating increased robustness and significantly lower costs, gradually taking a dominant position in the field. Wang et al.\cite{wang2017small} used the model trained by ImageNet Large Scale Visual Recognition Challenge (ILSVRC) data to complete the infrared small object detection task. Liangkui et al.\cite{liangkui2018using} combined with the data generated from oversampling, a multi-layer network was proposed for small object detection. Zhao et al.\cite{zhao2019tbc} developed an encoder-decoder detection method (TBC-Net) combining semantic constraint information of infrared small objects. Wang et al.\cite{wang2019miss} employed a generator and discriminator to address two distinct tasks: miss detection and false alarm, achieving a balance between these aspects. Nasser et al.\cite{nasrabadi2019deeptarget} proposed a deep convolutional neural network model for automatic object recognition (ATR). Zhang et al. proposed AGPCNet\cite{zhang2021agpcnet}, which introduced attention-guided context modules. Dai et al. introduced the asymmetric context modulation ACM\cite{dai2021asymmetric} and introduced the first real-world infrared small object dataset, SIRST. Wu et al.\cite{wu2022uiu} proposed a "U-Net within U-Net" framework to achieve multi-level representation learning of goals.\par
\section{Method}
In this section, we will be discussing HCF-Net in detail. As shown in Fig.~\ref{fig:domain_net}, HCF-Net is an upgraded U-Net architecture that consists of three crucial modules: PPA, DASI, and MDCR. These modules make our network more suitable for detecting small infrared objects and effectively tackle the challenges of small object loss and low background distinctiveness.
Next, we will provide a brief introduction to PPA in Sec.~\ref{sec:PPA}, followed by an overview of DASI in Sec.~\ref{sec:consis}, and finally, an introduction to MDCR in Sec.~\ref{sec:det}.
\subsection{Parallelized Patch-Aware Attention Module}
\label{sec:PPA}
In infrared small object detection tasks, small objects are prone to losing crucial information during multiple downsampling operations. As depicted in Fig.~\ref{fig:domain_net}, PPA substitutes traditional convolution operations in the encoder and decoder's fundamental components to better address this challenge.\par
\subsubsection{Multi-branch feature extraction}
The primary strength of PPA resides in its multi-branch feature extraction strategy. As depicted in Fig.~\ref{fig:PPA}, PPA employs a parallel multi-branch approach and Each branch is tasked with extracting features at various scales and levels. This multi-branch strategy facilitates the capture of multi-scale features of the object, consequently improving the accuracy of small object detection.
Specifically, this strategy involves three parallel branches: the local, global, and serial convolution branches. Given the input feature tensor $\mathbf{F}\in\mathbb{R}^{H'\times W'\times C}$, it is first adjusted through point-wise convolution to obtain $\mathbf{F}^{'}\in\mathbb{R}^{H'\times W'\times C'}$. Then, through the three branches, you can calculate $\mathbf{F}_{local}\in\mathbb{R}^{H'\times W'\times C'}$, $\mathbf{F}_{global}\in\mathbb{R}^{H'\times W'\times C'}$, and $\mathbf{F}_{conv}\in\mathbb{R}^{H'\times W'\times C'}$ separately. Finally, these three results are summed to obtain $\mathbf {\tilde{F}}\in\mathbb{R}^{H'\times W'\times C'}$.\par
Specifically, the distinction between the local and global branches is established by controlling the patch size parameter \( p \), which is realized through the aggregation and displacement of non-overlapping patches in spatial dimensions. Furthermore, we compute the attention matrix between non-overlapping patches to enable local and global feature extraction and interaction.\par
Initially, we employ computationally efficient operations, including Unfold and reshape, to partition \( \mathbf{F'} \) into a set of spatially contiguous patches (\( p \times p, H'/p, W'/p, C \)). Subsequently, we conduct channel-wise averaging to yield (\( p \times p, H'/p, W'/p \)), followed by linear computations using FFN~\cite{vaswani2017attention}. Subsequently, we apply the activation function to obtain the probability distribution in the spatial dimension for the linearly computed features and adjust their weights accordingly.\par
In the weighted outcomes, we employ feature selection~\cite{shi2023refocusing} to choose pertinent features for the task from tokens and channels. To be specific, let $d=\frac{H' \times W'}{p \times p}$, and represent the weighted outcome as $(\mathbf{t}_i)_{i=1}^{C'}$, where $\mathbf{t}_i \in \mathbb{R}^{d}$ represents the i-th output token. Feature selection operates on each token, yielding the output as $\hat{\mathbf{t}}_i = \mathbf{P} \cdot {sim}(\mathbf{t}_i,\mathbf{\xi}) \cdot \mathbf{t}_i$, where $\mathbf{\xi} \in \mathbb{R}^{C'}$ and $\mathbf{P} \in \mathbb{R}^{C' \times C'}$ are task-specific parameters, and ${sim}(\cdot,\cdot)$ is a cosine similarity function bounded within [0,1]. Here, $\mathbf{\xi}$ functions as the task embedding, specifying which tokens are relevant to the task. Each token $\mathbf{t}_i$ is reweighted based on its relevance to the task embedding (measured by cosine similarity), effectively simulating token selection. Subsequently, we apply a linear transformation of $\mathbf{P}$ for channel selection for each token, followed by reshape and interpolation operations, ultimately producing the features $\mathbf{F}_{{local}} \in \mathbb{R}^{H' \times W' \times C'}$ and $\mathbf{F}_{{global}} \in \mathbb{R}^{H' \times W' \times C'}$. Finally, we substitute the conventional 7x7, 5x5, and 3x3 convolution layers with a serial convolution consisting of three 3x3 convolution layers. This results in three distinct outputs: $\mathbf{F}_{conv1}\in\mathbb{R}^{H'\times W'\times C'}$, $\mathbf{F}_{conv2}\in\mathbb{R}^{H'\times W'\times C'}$, and $\mathbf{F}_{conv3}\in\mathbb{R}^{H'\times W'\times C'}$, which are then summed to obtain the serial convolution output $\mathbf{F}_{conv}\in\mathbb{R}^{H'\times W'\times C'}$.\par
\subsubsection{Feature fusion and attention}
Following feature extraction via the multi-branch feature extraction, we conduct adaptive feature enhancement using attention mechanisms.
The attention module comprises a sequence of efficient channel attention~\cite{wang2020eca} and spatial attention~\cite{woo2018cbam} 
components. In this context, $\mathbf{\tilde{F}} \in \mathbb{R}^{H \times W \times C'}$ is successively processed by a one-dimensional channel attention map $\mathbf{M}_c \in \mathbb{R}^{1 \times 1 \times C'}$ and a two-dimensional spatial attention map $\mathbf{M}_s \in \mathbb{R}^{H' \times W' \times 1}$. This process can be summarized as follows:
\begin{equation}
\begin{aligned}
\mathbf{F}_c = \mathbf{M}_c(\mathbf{\tilde{F}}) \otimes \mathbf{\tilde{F}},\hspace{0.2cm}\mathbf{F}_s = \mathbf{M}_s(\mathbf{F}_c) \otimes \mathbf{F}_c,
\end{aligned}
\label{eq:bce}
\end{equation}
\begin{equation}
\begin{aligned}
\mathbf{F^{''}} = \delta(\mathcal{B}({dropout}(\mathbf{F}_s))),
\end{aligned}
\label{eq:bce}
\end{equation}
where $\otimes$ denotes element-wise multiplication, $\mathbf{F}_c \in \mathbb{R}^{H \times W \times C'}$ and $\mathbf{F}_s \in \mathbb{R}^{H \times W \times C'}$ represent features after channel and spatial selection, $\delta(\cdot)$ and $\mathcal{B}(\cdot)$ represent Rectified Linear Unit (\textit{ReLU}) and Batch Normalization (\textit{BN}), respectively, and $\mathbf{F^{''}} \in \mathbb{R}^{H \times W \times C'}$ is the final output of PPA.
\subsection{Dimension-Aware Selective Integration Module}
\label{sec:consis}
\begin{figure}[h]
\begin{center}
\includegraphics[width=0.95\linewidth]{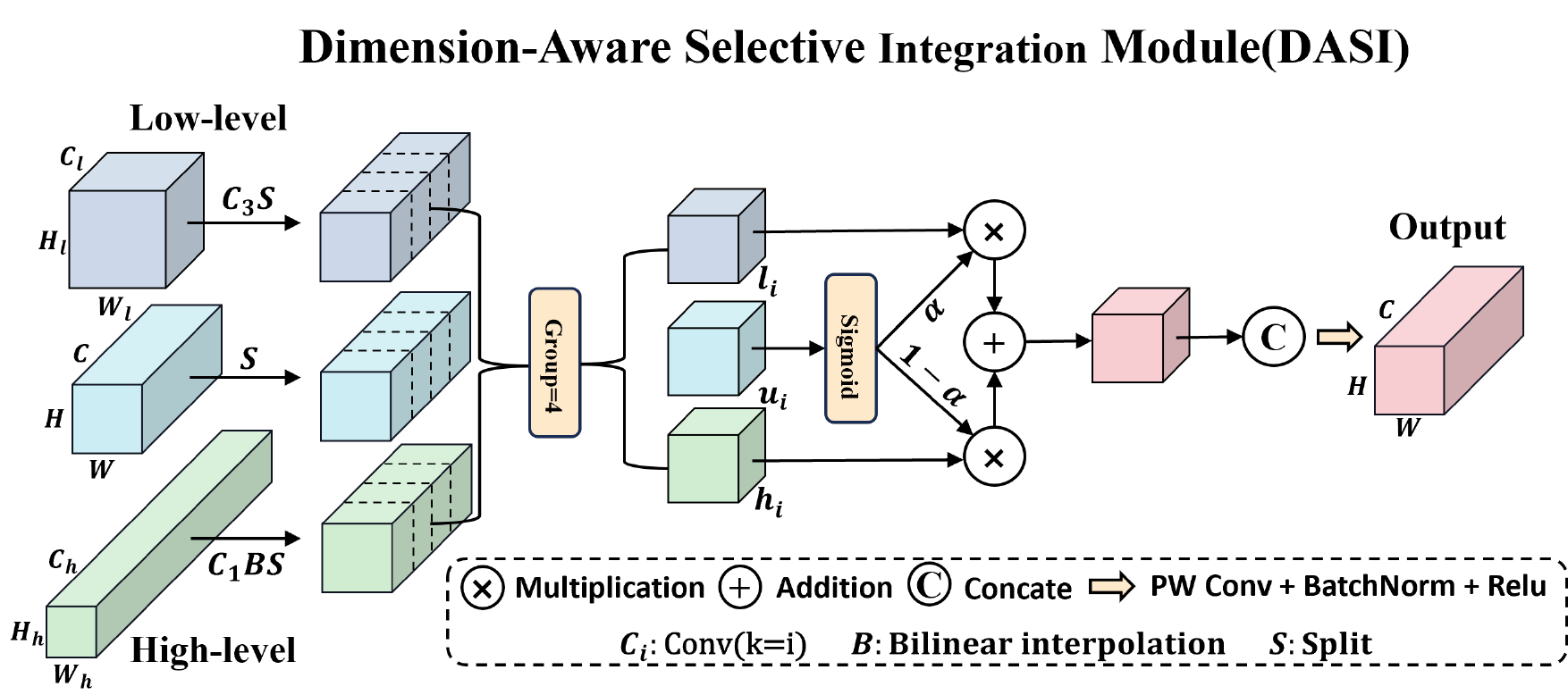}
\end{center}
   \caption{Detail structure of the dimension-aware selective integration module.}
\label{fig:DASI}
\end{figure}
During the multiple downsampling stages in infrared small object detection, high-dimensional features may lose information about small objects, while low-dimensional features may fail to provide sufficient context. To address this, we propose a novel channel partition selection mechanism (depicted in Fig.~\ref{fig:DASI}), enabling DASI to adaptively select appropriate features for fusion based on the object's size and characteristics.
In particular, DASI initially aligns the high-dimensional features $\mathbf{F_h} \in \mathbb{R}^{H_h \times W_h \times C_h}$ and low-dimensional features $\mathbf{F_l} \in \mathbb{R}^{H_l \times W_l \times C_l}$ with the features of the current layer $\mathbf{F_u} \in \mathbb{R}^{H \times W \times C}$ through operations like convolution and interpolation. Subsequently, it divides them into four equal segments in the channel dimension, resulting in $(\mathbf{h}_i)_{i=1}^{4} \in \mathbb{R}^{H \times W \times \frac{C}{4}}$, $(\mathbf{l}_i)_{i=1}^{4} \in \mathbb{R}^{H \times W \times \frac{C}{4}}$, and $(\mathbf{u}_i)_{i=1}^{4} \in \mathbb{R}^{H \times W \times \frac{C}{4}}$, where $\mathbf{h}_i$, $\mathbf{l}_i$, and $\mathbf{u}_i$ denote the i-th partitioned features of high-dimensional, low-dimensional, and current layer features, respectively. These partitions are computed according to the following formulas:
\begin{equation}
\begin{aligned}
\alpha={sigmoid}(\mathbf u_i),\hspace{0.2cm}\mathbf u_i^{'}=\alpha \mathbf l_i+(1-\alpha)\mathbf h_i,
\end{aligned}
\label{eq:bce}
\end{equation}
\begin{equation}
\begin{aligned}
\mathbf F_u'=[{\mathbf u_1',\mathbf u_2',\mathbf u_3',\mathbf u_4'}],\hspace{0.2cm}\hat{\mathbf{F_u}} = \delta\left(\mathcal{B}\left({Conv}(\mathbf{F_u'})\right)\right),
\end{aligned}
\label{eq:bce}
\end{equation}
where $\alpha \in \mathbb{R}^{H \times W \times \frac{C}{4}}$ represents the values obtained through the activation function applied to $\mathbf{u}_i$, $\mathbf{u}_i' \in \mathbb{R}^{H \times W \times \frac{C}{4}}$ represents the selectively aggregated results for each partition. After merging $(\mathbf{u}_i')_{i=1}^{4}$ in the channel dimension, we obtain $\mathbf{F_u'} \in \mathbb{R}^{H \times W \times C}$. The operations ${Conv}(·)$, $\mathcal{B}(·)$, and $\delta(·)$ denote convolution, batch normalization (\textit{BN}), and rectified linear unit (\textit{ReLU}), respectively, ultimately resulting in the output $\hat{\mathbf{F_u}} \in \mathbb{R}^{H \times W \times C}$.\par
If $\alpha> 0.5$, the model prioritizes fine-grained features, while if $\alpha< 0.5$, it emphasizes context features.\par
\subsection{Multi-Dilated Channel Refiner Module}
\label{sec:det}
In the MDCR, we introduce multiple depth-wise separable convolution layers with varying dilation rates to capture spatial features across a range of receptive field sizes, which allows for more detailed modeling of the differences between objects and backgrounds, enhancing its ability to discriminate small objects.\par
\begin{figure}[h]
\begin{center}
\includegraphics[width=0.95 \linewidth]{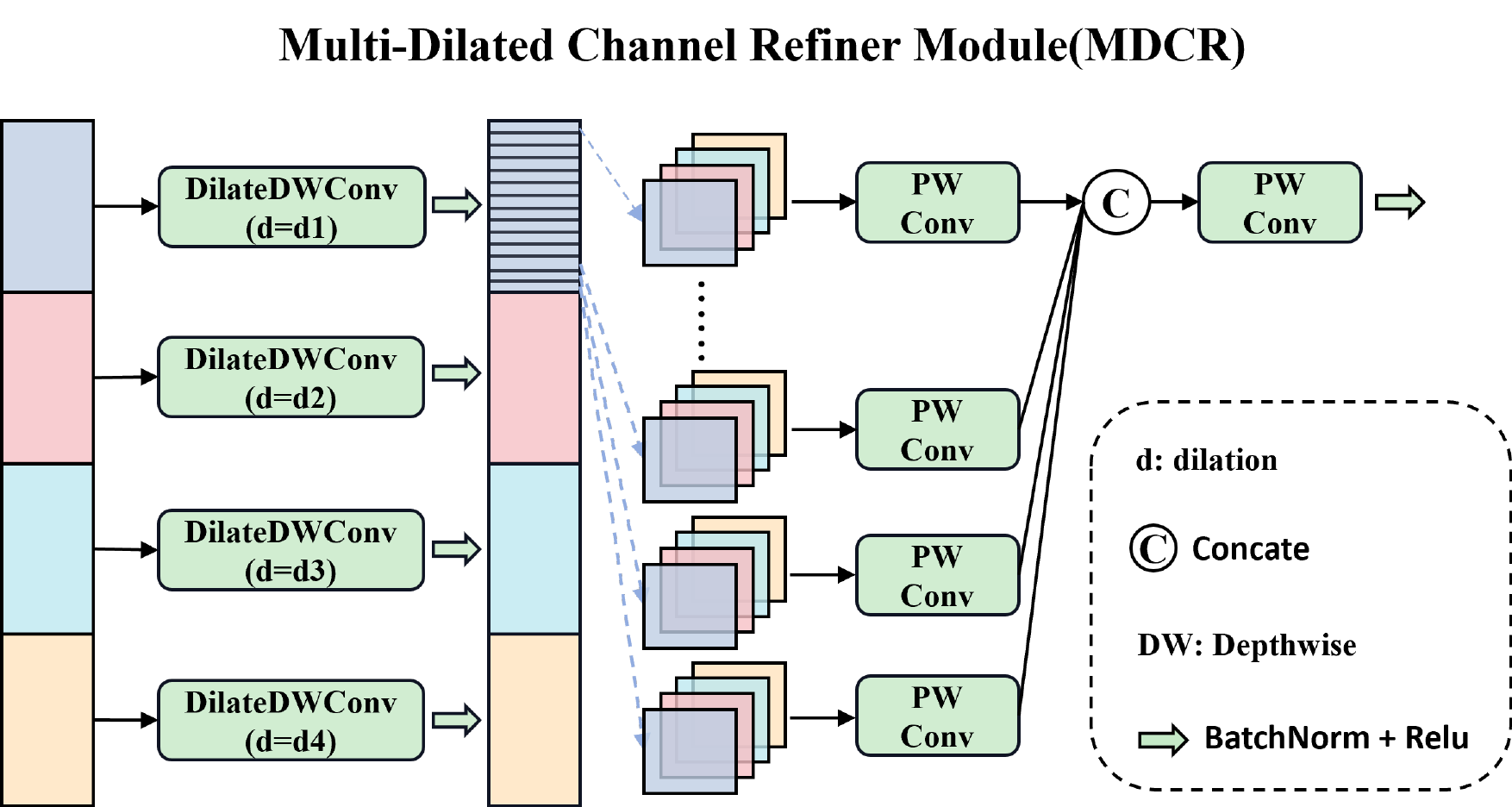}
\end{center}
   \caption{Detail structure of multi-dilated channel refiner module.}
\label{fig:MDCR}
\end{figure}

\begin{figure*}[ht]
\begin{center}
\includegraphics[width=0.85 \linewidth]{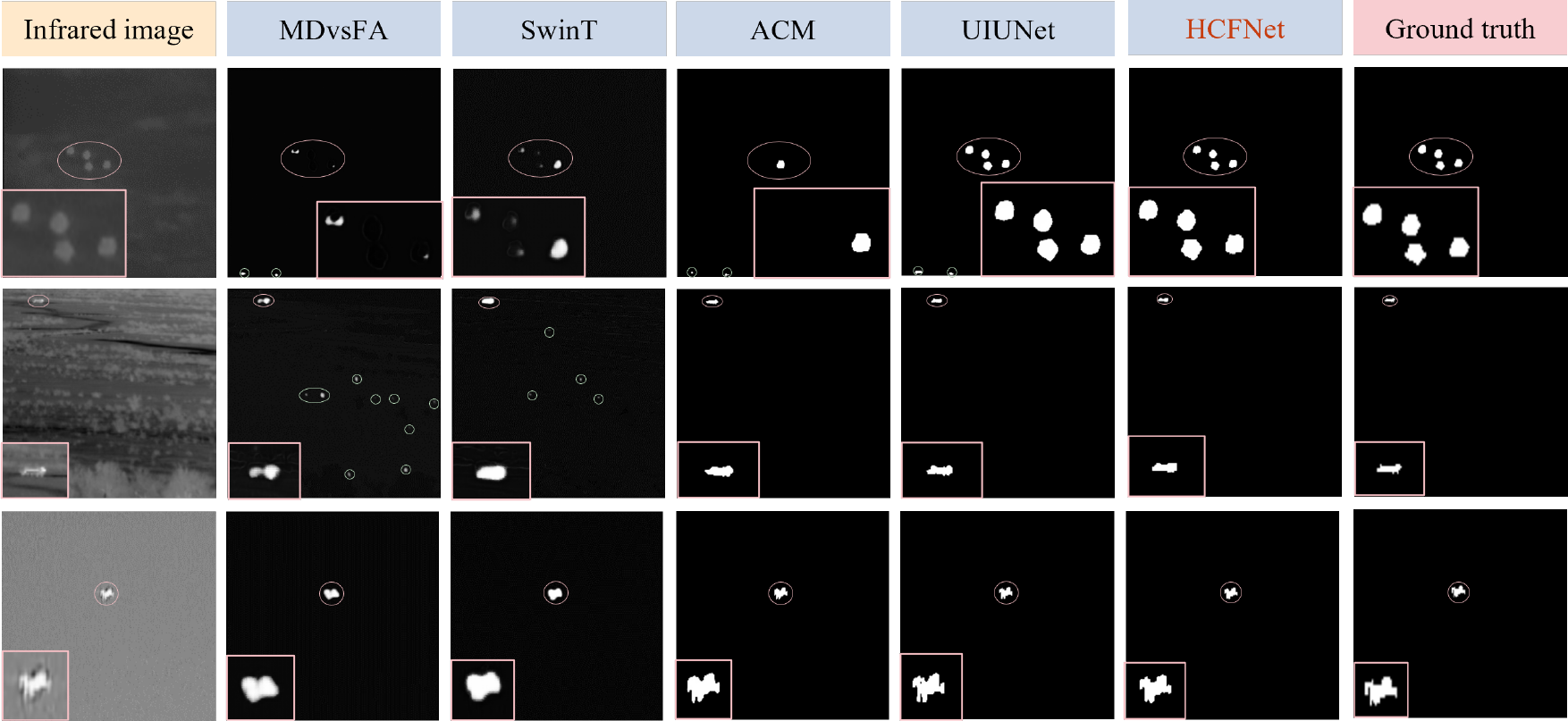}
\end{center}
   \caption{Visual examples of representative methods are provided. Pink and green circles represent true positive and false positive objects, respectively. The pink rectangle zooms in on true positive objects for a more apparent distinction of detection accuracy among different methods.
   \vspace{-0.5cm}
   }
\label{fig:visual}
\end{figure*}

Illustrated in Fig.~\ref{fig:MDCR}, MDCR partitions the input features $\mathbf{F_a} \in \mathbb{R}^{H \times W \times C}$ into four distinct heads along the channel dimension, generating $(\mathbf{a}_i)_{i=1}^{4} \in \mathbb{R}^{H \times W \times \frac{C}{4}}$. Each head then undergoes separate depth-wise separable dilated convolution with distinct dilation rates, yielding $(\mathbf{a}_i')_{i=1}^{4} \in \mathbb{R}^{H \times W \times \frac{C}{4}}$. We designate the convolution dilation rates as $d1$, $d2$, $d3$, and $d4$.
\begin{equation}
\begin{aligned}
\mathbf{a}_i' = DDWConv(\mathbf{a}_i),
\end{aligned}
\label{eq:bce}
\end{equation}
where $\mathbf{a}_i'$ denotes the features acquired by applying depth-wise separable dilated convolution to the $i$-th head. The operation $DDWConv(·)$ represents depth-wise separable dilated convolution, and $i$ takes values in ${1,2,3,4}$.\par
MDCR enhances the feature representation through channel segmentation and recombination. Specifically, we split $\mathbf{a}_i'$ into individual channels to obtain $(\mathbf{a}_i^j)_{j=1}^{\frac{C}{4}} \in \mathbb{R}^{H \times W \times 1}$ for each head. Following this, we interleave these channels across the heads to form $(\mathbf{h}_j)_{j=1}^{\frac{C}{4}} \in \mathbb{R}^{H \times W \times 4}$, thereby enhancing the diversity of multi-scale features. Subsequently, we perform inter-group and cross-group information fusion using pointwise convolution to obtain the output $\mathbf{F_o} \in \mathbb{R}^{H \times W \times C}$, achieving a lightweight and efficient aggregation effect.
\begin{equation}
\begin{aligned}
\mathbf{h}_j = \mathbf W_{inner}([\mathbf a_1^j, \mathbf a_2^j, \mathbf a_3^j, \mathbf a_4^j]),
\end{aligned}
\label{eq:bce}
\end{equation}
\begin{equation}
\begin{aligned}
\mathbf{F_o} = \delta(\mathcal{B}(\mathbf W_{outer}([\mathbf h_1, \mathbf h_2, ..., \mathbf h_j]))),
\end{aligned}
\label{eq:bce}
\end{equation}
where $\mathbf W_{inner}$ and $\mathbf W_{outer}$ are the weight matrices used in pointwise convolution. Here, $\mathbf{a}_i^j$ represents the $j$-th channel of the $i$-th head, while $\mathbf{h}_j$ denotes the $j$-th group of features. We have $i \in {1,2,3,4}$ and $j \in {1,2,...,\frac{C}{4}}$. The functions $\delta(·)$ and $\mathcal{B}(·)$ correspond to rectified linear units (\textit{ReLU}) and batch normalization (\textit{BN}), respectively.

\subsection{Loss design}
As depicted in Fig.\ref{fig:domain_net}, we employed a deep supervision strategy to further resolve the issue of small objects being lost during downsampling. The loss at each scale comprises binary cross-entropy loss and Intersection over union loss and is defined as follows:
\begin{equation}
\begin{aligned}
l_i = {Bce}(y, \hat{y}) + {Iou}(y, \hat{y}),\hspace{0.2cm}\mathcal{L} = \sum_{i=0}^{5} \lambda_i \cdot l_i,
\end{aligned}
\label{eq:bce}
\end{equation}
where $(l_i)_{i=0}^{5}$represents the losses at multiple scales, $\hat{y}$ is the ground truth mask, and $y$ is the predicted mask. The loss weights for each scale are defined as $[\lambda_0, \lambda_1, \lambda_2, \lambda_3, \lambda_4] = [1, 0.5, 0.25, 0.125, 0.0625]$.

\section{Experiments}

\begin{table}[!t]
\centering
\caption{Ablation study on the SIRST dataset in IoU(\%) and nIoU(\%). Here \CheckmarkBold means that this component is applied. Note that our baseline (Bas.).}
\label{tab:designs}
\setlength{\tabcolsep}{3pt}
\renewcommand\arraystretch{1}{
\begin{tabular}{clll|cc}  
\toprule
\multirow{2}{*}{\textbf{Bas.}} & \multicolumn{1}{c}{\multirow{2}{*}{\textbf{PPA}}} & \multicolumn{1}{c}{\multirow{2}{*}{\textbf{DASI}}} & \multicolumn{1}{c|}{\multirow{2}{*}{\textbf{MDCR}}} & \multicolumn{2}{c}{\textbf{SIRST}}   \\  
\cline{5-6}  
& \multicolumn{1}{c}{} & \multicolumn{1}{c}{} & & \textbf{IoU} & \textbf{nIoU} \\  
\hline
\CheckmarkBold & & & & 71.2 & 74.4 \\
\hdashline
\CheckmarkBold & \CheckmarkBold & & & 75.3 & 76.9 \\
\CheckmarkBold & \CheckmarkBold &\CheckmarkBold & & 77.9 & 76.1 \\
\CheckmarkBold & \CheckmarkBold & \CheckmarkBold & \CheckmarkBold & \textbf{80.1} & \textbf{78.3} \\
\hline
\end{tabular}}
\end{table}
\begin{table}[t]
\caption{Comparative evaluation on the SIRST dataset. We report metric IoU $(\%)$ and nIoU $(\%)$.}
\label{total results}
\centering
\renewcommand\arraystretch{1.1}{
\setlength{\tabcolsep}{6pt}{
\begin{tabular}{llllll}
\hline
\textbf{Method}    &  & \textbf{IoU}            & \textbf{nIoU}           \\ \cmidrule{1-4}
Top-Hat\cite{Zeng2006TheDO}$_{Infrared~Phys~Techn'2006}$   &                                   & 5.86           & 25.42          \\
LCM\cite{chen2013local}$_{T~Geosci~Remote'2013}$       &                                      & 6.84           & 8.96           \\
PSTNN\cite{PSTNN}$_{Remote~Sens-Basel'2019}$     &                                      & 39.44          & 47.72          \\
IPI\cite{Gao2013InfraredPM}$_{TIP'2013}$       &                                      & 40.48          & 50.95          \\
RIPT\cite{RIPT}$_{J-STARS'2017}$      &                                      & 25.49          & 33.01          \\
NIPPS\cite{dai2016infrared}$_{Infrared~Phys~Techn'2016}$     &                                      & 33.16          & 40.91          \\ 
MDvsFA\cite{wang2019miss}$_{ICCV'2019}$    & \multirow{7}{*}        & 56.17          & 59.84          \\
SwinT\cite{liu2021swin}~$_{ICCV'2021}$     &                                      & 70.53          & 69.89
\\
ACM\cite{dai2021asymmetric}$_{WACV'2021}$       &                                      & 72.45          & 72.15          \\
UIUNet\cite{wu2022uiu}~$_{TIP'2022}$ & & 78.25 & 75.15
\\ 
HCFNet (Ours) &                                      & \textbf{80.09}          & \textbf{78.31}
\\ \hline
\end{tabular}}
\vspace{-0.5cm}
}
\end{table}

\label{sec:exp}
\subsection{Datasets and Evaluation Metrics}
Our methods are assessed using SIRST\cite{dai2021asymmetric} in two standard metrics: Intersection over Union (IoU) and normalized Intersection over Union (nIoU)\cite{dai2021asymmetric}. SIRST was partitioned into training and test sets in an 8:2 ratio during our experiments.
\vspace{-0.3cm}
\subsection{Implementation Details.}
We perform experiments with HCF-Net on an NVIDIA GeForce GTX 3090 GPU. For input images of size 512×512 pixels and featuring three color channels, HCF-Net's computational cost is 93.16 GMac (Giga Multiply-Accumulate operations), comprising 15.29 million parameters. We employ the Adam optimizer for network optimization, employing a batch size of 4 and training the model 300 epochs.
\subsection{Ablation and Comparison}
\label{sec:study}
This section introduces ablative experiments and comparative experiments conducted on the SIRST dataset. Firstly, as shown in Table~\ref{tab:designs}, we use U-Net as a baseline and systematically introduce different modules to demonstrate their effectiveness. Secondly, as indicated in Table~\ref{total results}, our proposed method achieves outstanding performance on the SIRST dataset, with IoU and nIoU scores of 80.09\% and 78.31\%, respectively, significantly surpassing other methods. Finally, Fig.~\ref{fig:visual} presents visual results for various methods. In the first row, it can be observed that our method accurately detects more objects with a meager false-positive rate. The second row demonstrates that our method can still precisely locate objects in complex backgrounds. Finally, the last row indicates that our method provides a more detailed description of shape and texture features.\par

\section{conclusion}
In this paper, we address two challenges in infrared small object detection: small object loss and background clutter. To tackle these challenges, we propose HCF-Net, which incorporates multiple practical modules that significantly enhance small object detection performance. Extensive experiments have demonstrated the superiority of HCF-Net, outperforming traditional segmentation and deep learning models. This model is poised to be crucial in infrared small object detection.

\bibliographystyle{IEEEbib}
\bibliography{icme2024template}
\end{document}